\documentclass[10pt,twocolumn]{article} 
\usepackage{times}
\usepackage{graphicx}
\usepackage{amssymb}
\usepackage{url,hyperref}
\usepackage{color,soul}
\usepackage{adjustbox}

\usepackage{booktabs}
\makeatletter
\newcommand{\printfnsymbol}[1]{%
  \textsuperscript{\@fnsymbol{#1}}%
}
\makeatother

\date{}

\begin{document}

\title{3rd Place Solution to Google Landmark Recognition \\ Competition 2021}

\author{Cheng Xu\thanks{Equal contribution} , \quad Weimin Wang\footnotemark[1] , \quad Shuai Liu,\quad Yong Wang,\quad Yuxiang Tang, \\ \quad Tianling Bian,\quad Yanyu Yan,\quad Qi She,\quad Cheng Yang\\
ByteDance Inc.\\
\{xucheng.stephen, weimin.wang, liushuai.0530, wangyong.wy\}@bytedance.com\\
\{tangyuxiang, biantianling, yanyanyu, yangcheng.iron\}@bytedance.com\\
sheqi1991@gmail.com
}

\maketitle

\begin{abstract}
In this paper, we show our solution to the Google Landmark
Recognition 2021 Competition. Firstly, embeddings of images are extracted via various architectures (\emph{i.e.,} CNN-, Transformer- and hybrid-based), which are optimized by ArcFace loss. Then we apply an efficient pipeline to re-rank predictions by adjusting the retrieval score with classification logits and non-landmark distractors. Finally, the ensembled model scores $0.489$ on the private leaderboard, achieving 3rd place in the 2021 edition of the Google Landmark Recognition Competition. 

\end{abstract}

\section{Introduction}

Google Landmark Recognition 2021 Competition~\cite{link} is the
fourth landmark recognition competition on Kaggle, and this year it is organized together with ICCV 2021 Instance-Level Recognition workshop. Participants need to build models to recognize the landmarks (if any) correctly in a private test set, and the code submission method is adopted as previously. This year, the sponsor collects a new set of test images~\cite{DBLP:journals/corr/abs-2108-08874}, which
is created with a focus on fair worldwide representation. The training data for this competition comes from the Google Landmarks Dataset v2 (GLDv2)~\cite{weyand2020google}. GLDv2 is a large-scale benchmark for instance-level recognition and retrieval tasks, including approximately 5M images with about 200k distinct instance labels, which faces several challenges such as intra-class heterogeneity, class imbalance, and a large fraction of non-landmark test images. 
The cleaned subset of GLDv2 (GLDv2 CLEAN) consists of approximately 1.5 million images with 81,313 classes. Both GLDv2 and GLDv2 CLEAN can be used for training in this competition. Competition entries are evaluated using Global Average Precision (GAP) \cite{perronnin2009family, weyand2020google}. This paper summarizes our solution to the competition.

\begin{figure*}[t]
\begin{center}
\includegraphics[width=0.8\textwidth]{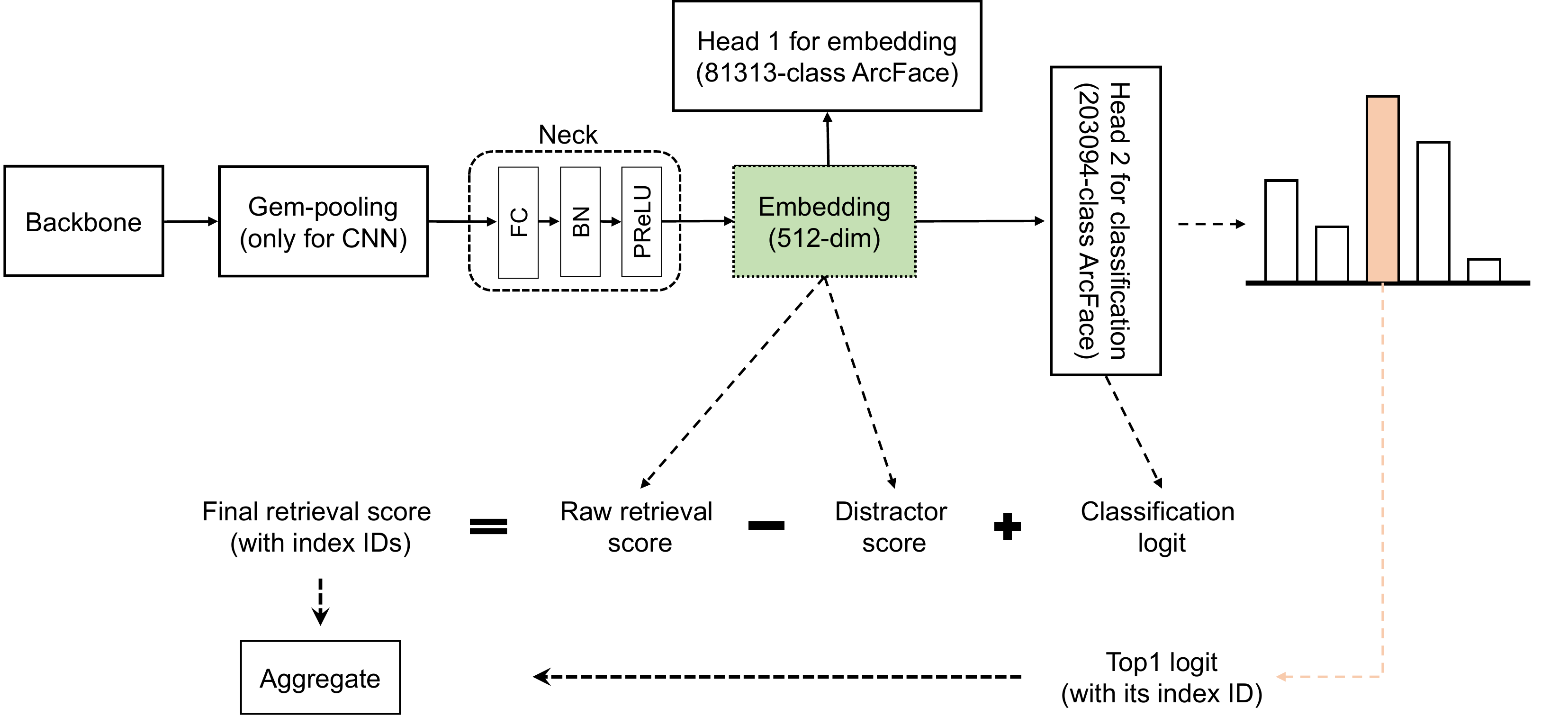}
\end{center}
\caption{The overview of the whole solution. The top diagram illustrates the model structure and the final prediction, aggregating retrieval score and top1 classification logit. Specifically, the retrieval score for each index ID is adjusted as: $Raw\ retrieval\ score$  - $Distractor\ score$ + $Classification\ Logit$ }
\label{fig1}
\end{figure*}

\section{Method}

\subsection{Overview}

Our final prediction comes from two parts: retrieval scores and classification logits. And the whole solution can be summarised as the following pipeline: 1) data prepossessing; 2) model training and retrieval (concat \& retrieval for model ensemble); 3) classification logits adjustment for class imbalance; 4) distractor score penalization for non-landmark; and 5) top 1 classification aggregation. The whole solution is shown in Figure~\ref{fig1}. Next, we will explain each part in detail.

\subsection{Data Prepossessing}
\label{sec:Data Prepossessing}
Follow previous solutions~\cite{DBLP:journals/corr/abs-2010-01650}, we split the training dataset as follows and we use landmark samples from 2019 test set as the validation set.
\begin{itemize}
  \item \textbf{GLDv2c}: the clean version of GLDv2, which consists of 1.5M images and 81313 landmarks.
  \item \textbf{GLDv2x}: all images from GLDV2 belong to 81313 landmarks, which consists of 3.2M images
  \item \textbf{GLDv2}: all images of GLDV2, which consists of 4.1M images and 203094 landmarks.
  \item \textbf{Non-landmark}: the non-landmark images from 2019 test set, which consists of 11k images.
\end{itemize}

\subsection{Model training and Retrieval}
\label{sec: Model training and Retrieval}
To calculate the similarity of different landmark samples, the 512-dimensional embeddings of input images are extracted from various backbone models.

\subsubsection{Model design}

As with previous solutions, the model architecture consists of backbone, gem pooling, neck for embedding and head for classification. To be specific, backbone outputs are aggregated via a Generalized-Mean (GeM) pooling layer~\cite{gu2018attention}, and then feed into an embedding neck layer(Linear(512)+1D-BN+PReLU). Finally, the image embeddings are used to classify specific landmarks supervised by ArcFace loss~\cite{deng2019arcface} with adaptive margin~\cite{ha2020google}. \emph{Note that the gem pooling is removed in transformer and hybrid-based models since each token output feature is the global representation because of the self-attention mechanism.}

Considering the diverse of model architecture, we chose three types of backbone, which as follows:
\begin{itemize}
  \item \textbf{CNN-based}: EfficientNet~\cite{tan2019efficientnet, tan2021efficientnetv2} B5, B6, B7, V2
  \item \textbf{Transformer-based}: Swin-L-384~\cite{liu2021Swin}
  \item \textbf{Hybrid-based}: CvT-W24-384~\cite{wu2021cvt}
\end{itemize}

Our final submission contains the above model backbones. Swin-L achieved the best retrieval performance $0.383$/$0.375$ on public/private, and other models perform as $B5<B6<CvT<B7<V2<Swin$ on retrieval score. It is believed that the greater the difference of model structure, the greater the complementarity of performance in the fusion stage. We found that Swin-L and CvT lead a more significant improvement of the final ensemble performance, compared with those CNN models.

\subsubsection{Training schedule}
Similar to last year's solutions, different image resolutions and training splits are adopted to accelerate convergence. Our training schedule could be divided into three stages.
\begin{itemize}
  \item \textbf{Stage 1}: GLDv2c is used to train the model to classify $81313$ landmarks. The model is pretrained by imagenet with $224$ input size.
  \item \textbf{Stage 2}: GLDv2x is used to finetuned the model and the weights from stage1 are used to pretrain. The resolution of input is $384$ or $448$ or $512$ for different models.
  \item \textbf{Stage 3}: GLDV2 is used to train the model to classify $203094$ landmarks for the classification logits. The weight from stage2 is utilized to pretrain. It is experimented that GLDV2 could not further improve the discrimination of embeddings. Thus, we freeze the backbone and neck, only optimize the classification head.
\end{itemize}
As for training details, each stage is trained for 10-20 epochs with a cosine annealing scheduler having one warm-up epoch. We use AdamW optimizer with learning rate of $[0.001,0.0001,0.00005]$ and weight decay of $0.05$ or $0.0001$. The batch-size varies between 512 and 1536 with Syc-BN on $[64,256]$ Tesla-T4-16GB gpus. For augmentation, RandAug~\cite{cubuk2020randaugment}, CutOut~\cite{devries2017improved} and RandomResizedCrop are adopted. With the improvement of image resolution, the setting of data augmentation increases gradually.

At inference time, we extract the features of the input image and retrieve the index set, then we select the $TopK$ ($K=7$) candidate images from the index set according to the retrieval similarity score, and we accumulate these candidates scores \emph{w.r.t.} their labels.

\subsection{Classification Logit Adjustment}
\label{sec:Classification}

We find that it is crucial to use classification logits to support predictions. More importantly, we found the top 1 pure classification accuracy of our B5 is around $0.39+$ on public LB, where for Swin-L it is around $0.34$. The clasification logit represents the cosine similarity between the image feature and the learned class-center by ArcFace. It is believed that this logit is complementary to retrieval score, especially when there are few images of some landmarks in the index.

At this stage, we get the Top 7 retrieved training images. So for each of the 7 images, we look up its classification logits from all 4 models chosen (B5 512 \& 768, B6 512 \& 768), and simply add the averaged logit to its corresponding retrieval cosine score as adjustment.

\subsection{Distractor Score penalization}
\label{sec:Distractor}

Similar to what~\cite{DBLP:journals/corr/abs-2010-01650} did last year, we use the 2019 test set's non-landmark images as index, and for each training image (4M), we find its Top 3 matched scores and simply take its average as the distractor score. Then, we generate a mapping between each 4M training image id to its distractor score, and upload this dict to Kaggle to use in submission.

We then subtract the distractor score for each adjusted cosine score above from Sec~\ref{sec:Classification}, and the final score for each retrieval is:

\textit{Raw retrieval score + classification logit - distractor score}

\subsection{Top 1 Classification Aggregation}
\label{sec:top1}

We find that using the top 1 classification logits from our best classification models (\textit{i.e.,} EfficientNet B5 and B6) can have another boost.

In Sec~\ref{sec:Distractor}, we obtain all Top 7 indexed images with their score adjusted, and we simply aggregate them $w.r.t.$ landmark id. However, in this section, we add another pair of (top 1 classification landmark id, top 1 classification logit) into the aggregation step. The classification logit used is just the raw top 1 logit, after we averaged all classification models' 200k prediction logits.

We can't penalize the classification pair as it is not from any image like the Top 7 matched, therefore it does not have a distractor score. But this turns out not to be a problem, as we found that top 1 classification score can be naturally used as penalty of non-landmark. 

\subsection{Ensembling}
\label{sec:ensembling}

For blending our various models (see Sec~\ref{sec: Model training and Retrieval}), we first l2-normalize each of them separately, concatenate them, and apply another l2-normalize to conduct ensembled retrieval. Next, we employ our ranking routine elaborated in Sec~\ref{sec:Classification} -~\ref{sec:top1} (the ensemble process is simply using the concatenated embedding space and running the whole procedure from start to finish on the larger embedding vectors)

\section{Conclusion}

In this paper, we presented our solution to the Google Landmark Recognition 2021 competition. We use features and classification logits extracted from several different models (CNN-, Transformer- and hybrid-based), optimized with an ArcFace Loss.
And we present an efficient re-ranking pipeline: retrieval, classification logit adjustment, distractor score adjustment and top 1 classification adjustment to generate more accurate landmark recognition results. After aggregating several models with different architectures, we reached a final score of $0.51829$ on the public and $0.48962$ on the private leaderboard respectively.

\bibliographystyle{abbrv}
\bibliography{main}
\end{document}